\documentclass[10pt]{article}

\usepackage[margin=0.78in]{geometry}
\usepackage{amsmath,amssymb,amsthm,mathtools}
\usepackage{booktabs}
\usepackage{graphicx}
\usepackage{float}
\usepackage{microtype}
\usepackage[numbers,sort&compress]{natbib}
\usepackage{xcolor}
\usepackage{hyperref}
\usepackage{enumitem}
\usepackage{multirow}
\usepackage{array}
\usepackage{tikz}
\usetikzlibrary{arrows.meta,positioning,fit}

\hypersetup{
  colorlinks=true,
  citecolor=blue!55!black,
  linkcolor=blue!55!black,
  urlcolor=blue!55!black,
  pdftitle={Paired Exact-Reset Evaluation of a Prediction-Derived Medium-to-Full World-Model Cascade},
  pdfauthor={Malo de Pastor}
}
\setlist{nosep,leftmargin=*}
\newtheorem{theorem}{Theorem}
\newtheorem{proposition}{Proposition}
\newtheorem{corollary}{Corollary}

\theoremstyle{definition}

\theoremstyle{remark}

\newcommand{\E}{\mathbb{E}}
\newcommand{\Prb}{\mathbb{P}}
\newcommand{\G}{\mathcal{G}}
\newcommand{\pos}[1]{\left(#1\right)_{+}}

\newcommand{\pusht}{PushT}
\newcommand{\pybullet}{PyBullet}

\title{\textbf{Paired Exact-Reset Evaluation of a}\\
Prediction-Derived Medium-to-Full World-Model Cascade}
\author{Malo de Pastor\\
\small Télécom SudParis, Institut Polytechnique de Paris, France\\
\small \href{mailto:depastorm@gmail.com}{\texttt{depastorm@gmail.com}}
\quad
\href{https://orcid.org/0009-0009-1913-0052}{ORCID: 0009-0009-1913-0052}}
\date{}

\begin{document}
\maketitle

\begin{abstract}
Existing adaptive-inference methods and recent world-action-model systems
already use cheap-stage outputs, intermediate states, or predicted futures to
allocate additional computation.  We study a narrower evaluation question:
under paired exact-reset physical outcomes, can a Medium-derived interface
predict when switching to a separately frozen Full predictor improves
task-specific decision loss enough to offset sequential overhead?  Our
contribution is a paired evaluation and audit protocol, rather than a new
generic routing rule: every candidate action is executed from the same reset
state, Medium and Full induce decisions over the same candidate set and
downstream task, and their paired physical-loss difference supplies the
post-hoc routing target.  On an initial fresh PushT audit bank (V106; 1,600
states, 39 tasks, three frozen checkpoint pairs), a frozen prediction-interface
router lowers overhead-inclusive decision cost relative to standalone Medium
by $0.00431$, standalone Full by $0.00258$, and a latency-advantaged task-only
router by $0.00262$.  We then use V106 only for development and prospectively
seal a second 1,600-state PushT confirmation (V107) against a stronger
current-state alternative: an input router using the task and a
dimension-matched projection of current DINO features plus all five candidate
actions, with no DINO encoder latency charged.  The prediction interface lowers
priced physical decision cost by $0.002549$ (state-clustered 95\% interval
$[-0.002867,-0.002238]$; one-sided 95\% upper bound $-0.002286$), and all three
fixed checkpoint-pair effects are negative.  A controlled-PyBullet audit
(12,000 states, 81 tasks, six regimes, three checkpoint pairs) independently
supports a composite task--prediction--regime router against fixed, margin, and
matched-random policies; a post-audit repair preserves shared-state pairing
without changing point estimates.  The sequential router remains slower than
fixed policies, and its fixed-policy advantage is restricted to low compute
prices.  The evidence supports incremental routing information in the tested
aligned prediction interface beyond one deliberately favoured current-DINO
control; it does not establish imagined-future causal sufficiency, compute
saving, a new adaptive-routing theory, closed-loop value, or cross-family
generality.
\end{abstract}

\section{Introduction}

Predictive models are useful because their outputs mediate decisions.
World models make this relationship unusually concrete: for a current
observation and candidate action, the model exposes an imagined consequence
that a controller may score before acting.  A natural response to difficult
queries is to spend more computation---use a larger predictor, draw more
rollouts, or run more denoising or planning steps.  Yet ``more predictive
compute'' is not an ordered intervention.  A higher-capacity model can improve
average predictive error while changing the ordering of the few actions that
matter, and the benefit can reverse across states and tasks.

Several concurrent systems already establish prediction- or
intermediate-state-conditioned world-model computation.  Gated geometric
Best-of-$N$ uses an initial action and predicted visual future to decide whether
to draw additional rollouts; SANTS learns where to stop video denoising from
intermediate video states; adaptive action execution uses prediction--reality
consistency to determine when another world-action-model pass is required; and
adaptive-depth latent world models route predictor depth during planning
\citep{zhao2026geometric,sun2026sants,wang2026imagination,
sivasankar2026adaptive}.  Generic adaptive neural cascades, post-hoc deferral,
and paid second-stage acquisition likewise learn whether the conditional
improvement of later computation exceeds its cost
\citep{bolukbasi2017adaptive,narasimhan2022posthoc,
jitkrittum2023confidence,regol2025acquisition}.  Accordingly, our novelty claim
does not concern the use of predicted futures for adaptive computation; it
concerns the paired physical estimand and frozen evaluation protocol used to
compare separately trained predictive capacities.  We ask:

\begin{quote}
\emph{Under paired exact-reset physical outcomes, can a Medium-derived
interface predict when switching to a separately frozen Full predictor
improves task-specific decision loss enough to offset sequential overhead?}
\end{quote}

We isolate this question through information interfaces.  A
prediction-interface router observes the downstream loss specification and
Medium's predicted consequence, but no true future, oracle regret, Full
prediction, or privileged simulator state.  Exact resets let us evaluate
Medium, Full, and all candidate actions on the same physical queries, making
the realized escalation benefit observable offline.  At test time the policy
is sequential: evaluate Medium, form its downstream action, and decide whether
to pay for Full.  A task-only control is allowed to route before Medium is
run, so it is latency-advantaged rather than an artificially weakened
sequential baseline.  Because that V106 control routes to Full on 99.1\% of
audit rows, we subsequently conduct a stronger prospective test.  V107 freezes
the prediction router and a dimension-matched input router that receives the
task, the current DINO representation, and every candidate action.  The
current-DINO control is deliberately favoured: its representation is treated as
already available, its routing decision precedes Medium, and no encoder latency
is charged.  Only after the comparison, thresholds, projection, bootstrap, and
fresh-state manifest are sealed do we generate a second 1,600-state exact-reset
bank.

The distinction between \emph{decision value} and \emph{compute saving} is
central.  We account for Medium, Full, task evaluation, and router overhead in
the objective.  The learned router reduces physical regret enough to beat both
fixed policies after pricing these latencies, but it remains slower in raw
milliseconds.  This result is useful---it identifies a learnable selective
benefit---while ruling out a stronger deployment claim.

Our contributions are:

\begin{enumerate}
  \item A candidate-complete paired exact-reset protocol for measuring
  task-specific physical decision losses of separately frozen Medium and Full
  predictors on identical state--task--action queries.
  \item A deployment-cost ledger that instantiates standard adaptive-cascade
  and value-of-information theory while explicitly comparing the Medium-first
  sequence with standalone Full.
  \item Frozen PushT and controlled-\pybullet{} audits, including a
  prospectively sealed second PushT bank against a latency-favoured,
  action-conditioned current-DINO control, state-clustered inference,
  compute-price sensitivity, and released paper-facing records.
\end{enumerate}

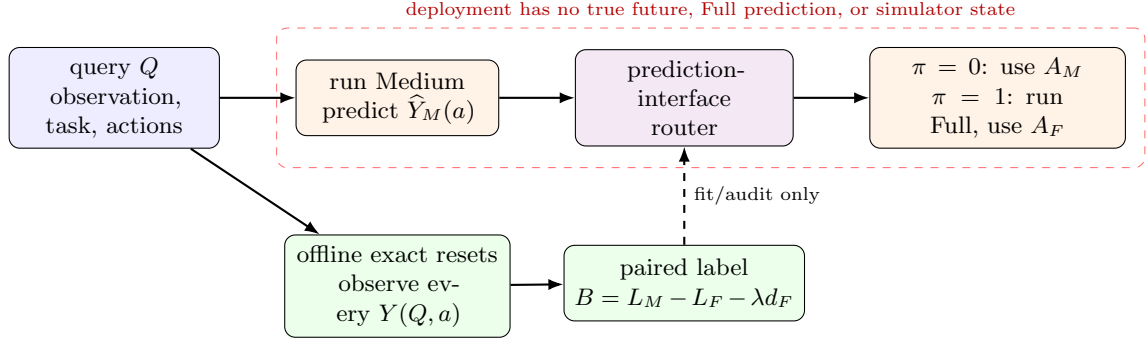
\begin{figure}[t]
\centering
\begin{tikzpicture}[
  node distance=10mm,
  box/.style={draw,rounded corners,align=center,inner sep=4pt,font=\small,
              minimum height=10mm},
  data/.style={box,fill=blue!7},
  compute/.style={box,fill=orange!10},
  decide/.style={box,fill=violet!9},
  audit/.style={box,fill=green!8},
  flow/.style={-{Latex[length=2mm]},thick},
  train/.style={-{Latex[length=2mm]},dashed,thick}
]
\node[data,text width=25mm] (q) {query $Q$\\observation, task, actions};
\node[compute,right=of q,text width=24mm] (m)
  {run Medium\\predict $\widehat Y_M(a)$};
\node[decide,right=of m,text width=26mm] (r)
  {prediction-interface\\router};
\node[compute,right=of r,text width=31mm] (out)
  {$\pi=0$: use $A_M$\\$\pi=1$: run Full, use $A_F$};
\draw[flow] (q) -- (m);
\draw[flow] (m) -- (r);
\draw[flow] (r) -- (out);
\node[audit,below=13mm of m,text width=27mm] (reset)
  {offline exact resets\\observe every $Y(Q,a)$};
\node[audit,below=13mm of r,text width=29mm] (label)
  {paired label\\$B=L_M-L_F-\lambda d_F$};
\draw[flow] (q) -- (reset);
\draw[flow] (reset) -- (label);
\draw[train] (label) -- node[right,font=\scriptsize] {fit/audit only} (r);
\node[draw=red!55,dashed,rounded corners,fit=(m)(r)(out),inner sep=2.5mm,
label={[font=\scriptsize,text=red!70!black]above:
deployment has no true future, Full prediction, or simulator state}] {};
\end{tikzpicture}
\caption{The deployed policy sees only the task and Medium's predicted
consequence.  Exact physical outcomes supervise and evaluate the router
offline, but are unavailable at routing time.}
\label{fig:interface}
\end{figure}

\section{Decision ledger for Medium-to-Full escalation}
\label{sec:theory}

\subsection{Queries and observable decision loss}

Work on a probability space supporting a physical query
$Q=(X,T,\mathcal A)$: current observation $X$, downstream task $T$, and finite
candidate action set $\mathcal A$.  All population statements below are
conditional on a known frozen Medium/Full checkpoint pair $r$, with its
associated router, threshold, and latency constants; we suppress $r$ in the
notation.  Equivalently, when averaging the finite three-checkpoint design,
the checkpoint-pair index is included in every compared information field,
and the reported risk is the prespecified equally weighted average of the
three checkpoint-specific risks.  A frozen model $m\in\{M,F\}$ predicts
$\widehat Y_m(Q,a)$ and selects
\[
 A_m(Q)\in\arg\min_{a\in\mathcal A}
 \ell_T(\widehat Y_m(Q,a)),
\]
using a fixed tie rule.  Exact reset exposes $Y(Q,a)$ for every candidate.
We use candidate-set physical regret
\[
 L_m(Q)=\ell_T(Y(Q,A_m))
 -\min_{a\in\mathcal A}\ell_T(Y(Q,a)).
\]
Thus $L_M$ and $L_F$ are observed on the same query; no claim is made about
actions outside $\mathcal A$ or real-robot causal effects.  Pairing is
source-enforced for every candidate: PushT resets to the recorded seeded base
state before each rollout, and PyBullet restores a saved simulator state.
Deterministic duplicate replay is additionally checked on three states per
PushT shard and one candidate per checked state; this is sampled validation,
not an exhaustive replay test over all states and actions.  Model action ties
use the frozen deterministic tie rule.  V106 router scores are replayed as
NumPy float32 and escalate only under the strict comparison
$\text{score}>\text{threshold}$; the V104D/V104E audit environment pins the
serialization runtime and records archived escalation-mask checksums.

\subsection{Sequential and standalone costs}

Let $s_{\G}(Q)$ be the always-paid latency for running Medium and exposing
interface $\G$, and let $d_F(Q)$ be the additional latency of evaluating and
scoring Full after Medium.  The general notation also permits query-dependent
standalone latencies $c_M(Q)$ and $c_F(Q)$.  In the executed analyses these
latencies are frozen constants within checkpoint pair and latency summary, so
expectations average them only across the prespecified checkpoint weights.
Assume the relevant losses and costs are integrable.  At compute price $\lambda\geq0$, define realized net escalation
benefit
\begin{equation}
  B=L_M-L_F-\lambda d_F .
  \label{eq:benefit}
\end{equation}
Positive $B$ means escalation reduces the priced decision objective.

A sequential interface $\G$ is the sigma-field available after Medium.  For
example,
\[
  \G_0=\sigma(T,R),\qquad
  \G_1=\sigma(T,R,S_M),
\]
where $T$ is the downstream task, $R$ denotes any information held fixed
between the interfaces (such as a prespecified regime descriptor), and $S_M$
is Medium's predicted physical consequence.  A $\G$-measurable
binary policy $\pi$ has sequential cost
\begin{equation}
  C_{\G}(\pi)=L_M+\lambda s_{\G}-\pi B.
  \label{eq:routercost}
\end{equation}
Define $b_{\G}=\E[B\mid\G]$.

\begin{proposition}[Standard Bayes escalation rule and available value]
\label{prop:bayes}
An optimal $\G$-measurable rule is
\[
  \pi^\star_{\G}=\mathbf{1}\{b_{\G}>0\},
\]
up to ties.  Its expected cost is
\[
  \E[C_{\G}(\pi^\star_{\G})]
  =\E[L_M+\lambda s_{\G}]-V(\G),
  \qquad
  V(\G):=\E[\pos{b_{\G}}].
\]
\end{proposition}

Thus $V(\G)$ is the improvement available from conditional escalation after
the incremental Full cost has been priced, but before comparing the
always-paid sequential latency to a standalone policy.

\begin{proposition}[Sequential-to-standalone accounting identity]
\label{prop:standalone}
Let $c_M$ and $c_F$ be the measured standalone latencies of Fixed Medium and
Fixed Full, and define the signed sequential-to-standalone Full latency gap
$\kappa_{\G}=s_{\G}+d_F-c_F$.  In the evaluated protocols this gap is
positive, so it is a duplication penalty; the identity itself does not require
nonnegativity.  The Bayes sequential router's expected gains
over the standalone policies are
\begin{align}
 \E[C_M^{\mathrm{fix}}-C_{\G}(\pi^\star_{\G})]
 &=V(\G)-\lambda\E[s_{\G}-c_M],
 \label{eq:gainmedium}\\
 \E[C_F^{\mathrm{fix}}-C_{\G}(\pi^\star_{\G})]
 &=\E[\pos{-b_{\G}}]-\lambda\E[\kappa_{\G}],
 \label{eq:gainfull}
\end{align}
where $C_m^{\mathrm{fix}}=L_m+\lambda c_m$.
\end{proposition}

Equation~\eqref{eq:gainfull} prevents a common accounting error.  The
negative-part term is the value of not escalating on unfavorable queries, but
standalone Full avoids the Medium pass entirely.  A sequential router beats
Fixed Full only when its optionality exceeds the priced signed latency gap.
In both evaluated protocols $\kappa_{\G}>0$, so the gap is a duplication
penalty; our routers overcome it at the frozen price despite higher raw
latency.

\subsection{Value of the predicted consequence}

Suppose $\G_0=\sigma(T,R)\subseteq\G_1=\sigma(T,R,S_M)$, so that the only
added source is Medium's predicted consequence.

\begin{theorem}[Nested-interface Jensen identity]
\label{thm:piv}
Define
\[
 \operatorname{PIV}(\G_1;\G_0)
 :=V(\G_1)-V(\G_0).
\]
Then
\begin{equation}
 \operatorname{PIV}(\G_1;\G_0)
 =
 \E\!\left[
   \pos{\E[B\mid\G_1]}-\pos{\E[B\mid\G_0]}
 \right]\geq0.
 \label{eq:piv}
\end{equation}
Writing $U=\E[B\mid\G_1]$, the gap is strictly positive exactly when a
positive-probability set of coarse outcomes has both
$\E[U_+\mid\G_0]>0$ and $\E[(-U)_+\mid\G_0]>0$.  For a common incremental Full cost, the net deployment value of
replacing $\G_0$ by $\G_1$ is
\[
 \operatorname{PIV}(\G_1;\G_0)
 -\lambda\E[s_{\G_1}-s_{\G_0}].
\]
\end{theorem}

The theorem is a population information statement, not a guarantee that an
arbitrary finite-sample learner will exploit the interface.  Moreover, our
sealed \pusht{} task-only control routes before Medium and is therefore
latency-advantaged.  It is a strict deployment comparator under frozen
function classes and data, not a plug-in estimator of
Eq.~\eqref{eq:piv}.

\subsection{Finite-sample selection}

Conditional on all development data, validation choices, fitted parameters,
and algorithmic randomness, let $\widehat b$ be a $\G$-measurable score and
let $\widehat\pi=\mathbf{1}\{\widehat b>0\}$.  Any executed strict
threshold rule $\mathbf 1\{s>\tau\}$ has this form by taking
$\widehat b=s-\tau$.  Assume $\E|b_{\G}|<\infty$; for the finite upper bound below also assume
$\E|\widehat b-b_{\G}|<\infty$.

\begin{proposition}[Standard plug-in sign-regret identity]
\label{prop:plugin}
For a fixed interface,
\begin{equation}
  \E[C_{\G}(\widehat\pi)]
  -\E[C_{\G}(\pi^\star_{\G})]
  =
  \E\!\left[
    |b_{\G}|\,
    \mathbf{1}\{\widehat\pi\neq\pi^\star_{\G}\}
  \right]
  \leq \E[|\widehat b-b_{\G}|].
  \label{eq:plugin}
\end{equation}
\end{proposition}

This weighted sign identity explains why noisy individual benefit prediction
can still improve aggregate decisions: errors matter only when they cross
zero, and their decision penalty is the true conditional margin.

\begin{corollary}[Learned-policy deployment decomposition]
\label{cor:learned}
Let
\[
 \mathcal R_{\G}(\widehat\pi)
 :=\E\!\left[
 |b_{\G}|\mathbf 1\{\widehat\pi\ne\pi^\star_{\G}\}
 \right].
\]
The learned router's gains over standalone Medium and Full equal,
respectively,
\begin{align}
 V(\G)-\lambda\E[s_{\G}-c_M]-\mathcal R_{\G}(\widehat\pi),
 \label{eq:learnedmedium}\\
 \E[\pos{-b_{\G}}]-\lambda\E[\kappa_{\G}]
 -\mathcal R_{\G}(\widehat\pi).
 \label{eq:learnedfull}
\end{align}
\end{corollary}

Equations~\eqref{eq:learnedmedium}--\eqref{eq:learnedfull} separate the three
ways a sequential method can fail: insufficient conditional optionality,
excess acquisition/duplication cost, or imperfect selection from the available
interface.

\begin{corollary}[Margin corollary]
\label{cor:margin}
If $A<\infty$, $\alpha\geq0$, $\varepsilon\geq0$,
$\Prb(|b_{\G}|\le t)\le At^\alpha$ for all $t>0$, and
$|\widehat b-b_{\G}|\le\varepsilon$ almost surely, then the excess cost in
Eq.~\eqref{eq:plugin} is at most $A\varepsilon^{\alpha+1}$.
\end{corollary}

\begin{proposition}[Scalar-ordering obstruction]
\label{prop:threshold}
Let $Z$ be a real-valued $\G$-measurable score and let the high-score
threshold policy be $\pi_\tau=\mathbf 1\{Z>\tau\}$.
Suppose there are $\G$-measurable sets $A_+$ and $A_-$ with positive
probability and a $\gamma>0$ such that
\[
 b_{\G}\geq\gamma\ \text{on }A_+,\qquad
 b_{\G}\leq-\gamma\ \text{on }A_-,
\]
while
\[
 \operatorname*{ess\,sup}_{A_+} Z
 <
 \operatorname*{ess\,inf}_{A_-} Z.
\]
Then every high-score threshold has Bayes excess cost at least
\[
 \gamma\min\{\Prb(A_+),\Prb(A_-)\}.
\]
\end{proposition}

Thus ``escalate when difficulty is high'' requires a single-crossing
relationship between the score and conditional benefit.  A score that ranks a
negative-benefit region above a positive-benefit region has irreducible
threshold regret.  The matched Medium-margin baseline probes this issue
empirically, but its failure alone does not verify the proposition's setwise
assumptions.

\begin{proposition}[Fixed-rate rearrangement lemma]
\label{prop:matched}
Fix $q\in[0,1]$ and allow a $\G$-measurable routing propensity
$g:\Omega\to[0,1]$ with $\E[g]=q$.  On an enlarged probability space
carrying an independent $U\sim\operatorname{Unif}(0,1)$, the realized binary
action $\pi=\mathbf 1\{U\le g\}$ is measurable with respect to
$\G\vee\sigma(U)$ and satisfies $\E[\pi\mid\G]=g$.  Among these propensities, an optimizer is
\[
 g^\star=\mathbf 1\{b_{\G}>t\}
 +\rho\,\mathbf 1\{b_{\G}=t\},
\]
where $t$ and $\rho\in[0,1]$ are chosen so that $\E[g^\star]=q$; the
endpoint cases $q=0$ and $q=1$ are constant propensities.  This upper-tail rule
maximizes $\E[b_{\G}g]$.  Its advantage over independent rate-$q$ routing is
\[
 \E[b_{\G}g^\star]-q\E[B]\ge0.
\]
For $0<q<1$, the inequality is strict if and only if $b_{\G}$ is not almost
surely constant.
\end{proposition}

This isolates selection from escalation count.  Our random controls are
input-routing controls and hence avoid Medium on Full-routed queries; their
lower latency makes the empirical comparison conservative rather than an
exact evaluation of Proposition~\ref{prop:matched}.

\begin{table}[t]
\centering
\caption{Theory-to-evidence map.  The algebra organizes the experiment; the
new evidence lies in paired protocols and frozen audit comparisons.}
\label{tab:theorymap}
\small
\begin{tabular}{p{0.27\linewidth}p{0.25\linewidth}p{0.40\linewidth}}
\toprule
Object & Status & Empirical role \\
\midrule
Bayes escalation/value & Standard, adapted & Defines the priced escalation target and oracle. \\
Nested-interface Jensen gap & Standard value of information & Defines population prediction-interface value. \\
Standalone-Full decomposition & Protocol-specific identity & Separates optionality from the signed sequential-to-standalone Full latency gap. \\
Plug-in sign regret & Standard, adapted & Explains why noisy benefit scores can still select usefully. \\
Finite-audit observability & Protocol-specific formalization & A complete executed record makes candidate-level costs computable; the compact release recomputes paper-facing statistics. \\
Frozen audit policy gains & New empirical evidence & Tested in V104E, V106, and the prospective V107 current-DINO/action confirmation. \\
\bottomrule
\end{tabular}

\end{table}

\section{Paired physical routing}
\label{sec:method}

\subsection{Common protocol}

For every current physical state, we construct the same candidate action set
for Medium and Full.  Exact resets provide a paired record
\[
 \bigl\{
   S_M(Q,a),S_F(Q,a),Y(Q,a)
 \bigr\}_{a\in\mathcal A},
\]
where $S_m$ is a predicted consequence and $Y$ the realized physical
consequence.  Each downstream task maps predicted consequences to candidate
scores, selects an action, and evaluates its physical regret using $Y$.
Because Medium and Full are observed on the same query and action set, their
realized loss difference is paired rather than confounded by state sampling.

\begin{proposition}[Finite-record recomputability and target-law observation]
\label{prop:paired}
Suppose a complete executed audit record contains every candidate physical
outcome, all frozen model decisions and policy inputs, and all randomness
needed to replay stochastic components.  Then the realized cost of any frozen
policy and every paired candidate-minus-reference difference are measurable
functions of that record; their finite-audit empirical means are exactly
computable.  The public compact release provides the weaker property needed
for the reported audit: it contains policy-facing per-state/per-task arrays
sufficient to recompute every paper-facing statistic, but not every candidate
physical outcome or upstream prediction from first principles.

A target-population claim requires a separate sampling model.  Let
physical-state clusters be sampled from a declared law or known sampling
design, and let task, bank, and checkpoint-pair weights be fixed in advance.
The resulting target mean is a functional of the observable cluster law.
Primary confidence statements are model-based over physical-state clusters:
conditional on the fixed tasks, bank weights, and three frozen checkpoint
pairs, generated states within each declared simulator-state generator or bank
are treated as iid draws from that law.  The fixed audit mean is exactly known
without this assumption; its confidence interpretation depends on it, and the
chosen reset seeds are not a randomized-treatment design.  Under iid cluster
sampling, or a design-valid analogue, and finite second moments, the sample
mean is consistent and admits standard mean asymptotics.  Bootstrap confidence
statements require additional validity conditions and a resampling scheme that
preserves every clustered and crossed factor.  Treating checkpoint pairs as
random additionally requires a declared checkpoint-generating population;
three fixed checkpoints alone define only a finite equally weighted
checkpoint estimand.
\end{proposition}

This separates deterministic recomputability from population inference.
Neither property turns the finite candidate-set simulator intervention into an
unrestricted robotics or real-world causal-generalization result.

For checkpoint pair $r$, the executed V106 router regresses the paired gross
regret difference $\Delta_r=L_{M,r}-L_{F,r}$.  Because measured incremental
Full latency is constant across queries within that checkpoint pair,
$d_{F,r}(Q)=d_{F,r}$, the Bayes net-benefit score satisfies
$\E[B_r\mid\G_r]=\E[\Delta_r\mid\G_r]-\lambda d_{F,r}$.  Thus an exactly
calibrated gross conditional-mean score could be converted to the Bayes net
rule by thresholding at $\lambda d_{F,r}$.  The executed fitted score is not
assumed perfectly calibrated: for each checkpoint pair, interface, and compute
price, a strict threshold is selected from a validation grid consisting of
the two constant endpoints and 101 empirical score quantiles.  This is a
frozen threshold-policy implementation, not an estimate of the Bayes
net-benefit function.  If incremental Full latency varies by query, the Bayes rule depends on
$\E[\Delta_r-\lambda d_{F,r}\mid\G_r]
=\E[\Delta_r\mid\G_r]-\lambda\E[d_{F,r}\mid\G_r]$.
A constant threshold on a gross-benefit score is Bayes-equivalent only when
$\E[d_{F,r}\mid\G_r]$ is constant within the deployed checkpoint pair, or
when its conditional variation is already absorbed into the score.  Observing
realized latency at routing time is sufficient but not necessary.  The recovered V104D source
uses the same gross target, fits routers only on training rows, and selects
strict thresholds only on validation rows.  V104E applies the frozen V104D
router and thresholds without refitting on confirmation outcomes.  Both routers use a two-hidden-layer MLP (128 and 64 ReLU units).  The \pusht{} candidate uses the task
description and Medium's calibrated predicted consequence.  The \pybullet{}
candidate additionally uses a prespecified regime descriptor.  Neither uses
the explicit simulator state, true future, realized losses, oracle action,
Full prediction, or correctness labels.

At audit time, Medium is always evaluated.  The router observes its frozen
interface and either retains Medium's action or evaluates Full and uses Full's
action.  The reported objective is
\[
  \text{decision regret}+\lambda\times\text{measured latency (ms)},
\]
with $\lambda=0.002$.  Candidate-minus-reference differences are negative when
the candidate is better.  Sequential latency follows
$s_{\G}+\pi d_F$: a Full escalation adds a complete Full prediction/decision
pass to the already-paid Medium interface.

\subsection{Prospective prediction-versus-current-DINO confirmation}
\label{sec:v107method}

The initial V106 audit is development data for the stronger V107 control; none
of its rows are reused as confirmatory evidence.  Before any V107 physical
outcome is generated, we seal one primary comparison:
\[
  C_{\mathrm{prediction}}-C_{\mathrm{DINO+actions}}.
\]
Both interfaces use the same checkpoint-pair-specific
\texttt{StandardScaler+MLPRegressor} family and regress the paired gross
Medium-minus-Full physical regret difference.  The prediction interface has 22
task features plus the complete 43-dimensional Medium-derived consequence
interface.  The control has the same 22 task features plus a 43-dimensional
projection of the current 64-D DINO representation concatenated with the five
candidate actions.  Its scaler and PCA are fit only on V106 training states.
Thus both have 65 input dimensions, while the control receives direct action
information and routes before Medium.  We treat its current DINO representation
as already available and charge no DINO encoder latency.

The six serialized direct-gross routers, checkpoint-specific thresholds,
float32 score semantics, latency ledger, projection, 39 tasks, five actions,
three checkpoint pairs, 1,600 fresh reset seeds, and audit code are hashed in
the confirmation manifest before generation.  Thresholds are selected only on
V106 validation states.  V107 uses two new 800-state shards and no fitting,
threshold selection, debugging, or scientific design change may use their
outcomes.  The single primary gate averages tasks and the three fixed
checkpoint-pair effects within physical state, resamples 1,600 states for
20,000 bootstrap repetitions, and passes only if the aggregate point and
one-sided 95\% upper bound are negative and every checkpoint-pair point is
negative.  Secondary price and fixed-policy displays cannot rescue or redefine
this gate.

\subsection{References and inference}

The primary references are:

\begin{itemize}
  \item \textbf{Fixed Medium} and \textbf{Fixed Full};
  \item a \textbf{task-only input router} in \pusht{}, with the same MLP family
  but no state- or prediction-dependent signal;
  \item a cheaper \textbf{Medium-margin sequential router} at the candidate's
  escalation rate;
  \item \textbf{matched random input routing}, preserving the candidate rate
  (and exact seedwise counts in \pusht{}).
\end{itemize}

We cluster paired inference at physical state and average tasks within state.
V107 has one primary comparison and uses the prospective gate in
Section~\ref{sec:v107method}; its ordinary 95\% interval and one-sided 95\%
upper bound are conditional on the fixed tasks and checkpoint pairs.  V106
averages the three fixed seed effects for each shared state before
resampling 1,600 states.  Its post-audit crossed seed sensitivity resamples
three seed positions and one common physical-state index vector shared across
all sampled seed positions; it remains secondary and does not establish
population coverage over training randomness.
The originally executed V104E bootstrap resampled different state-index
vectors inside sampled seed--bank cells and therefore failed to preserve the
crossed physical-state pairing.  After external audit, we repaired only the
uncertainty calculation: the primary analysis averages the three frozen seed
effects within each of 12,000 physical states, resamples states within each of
six fixed equally weighted banks, and uses 10,000 percentile replicates.  A
crossed sensitivity resamples model seeds and one common state-index vector per
bank, shared across sampled seeds.  Point estimates, policies, outcomes, and
the four-reference family are unchanged.  Bonferroni-adjusted percentile-bootstrap upper quantiles are $0.99$ for V106
and $0.9875$ for V104E; every corrected primary and crossed-sensitivity upper
bound remains below zero.  These are approximate bootstrap statements
conditional on the fixed tasks, banks, checkpoints, and cluster-sampling
interpretation, not exact finite-sample family-wise guarantees.  Random input policies and
the PushT task-only policy avoid duplicated Medium computation, so the
candidate must overcome their latency advantage.

\section{Frozen audit evaluations}
\label{sec:experiments}

\begin{table}[t]
  \centering
  \caption{The frozen audits share the routing question.  V107 is a second
  prospective PushT bank against a stronger current-DINO/action control;
  PyBullet covers a different physical mechanism.  The predictors remain
  related DINO-style families rather than distinct world-model paradigms.}
  \label{tab:env}
  \small
  \begin{tabular}{lrrrr}
\toprule
Evaluation & States & Tasks & Model seeds & Primary references \\
\midrule
PushT V106 (initial fresh bank) & 1,600 & 39 & 3 & 5 \\
PushT V107 (prospective control) & 1,600 & 39 & 3 & 1 \\
PyBullet V104E (6 regimes) & 12,000 & 81 & 3 & 4 \\
\bottomrule
\end{tabular}

\end{table}

\paragraph{\pusht{}.}
We generate ten exact-reset shards from the public \pusht{} simulator.  Shards
0--7 supply 5,200 training and 1,200 validation states; shards 8--9 are a
sealed audit set of 1,600 fresh states and are never opened during router or
threshold selection.  Each query has five candidate actions and horizon 30.
The task family crosses 13 block-displacement targets with balanced,
position-weighted, and orientation-weighted losses (39 tasks).  Medium and
Full are frozen residual MLP predictors over a 64-D PCA of pooled DINOv2
ViT-S/14 features.  Medium has two 256-wide GELU layers (99,392 parameters);
Full has three 512-wide layers (592,448 parameters).  Both were trained with
AdamW on 4,950 action-conditioned rows, selected on 900 validation rows, and
then passed through capacity-specific ridge physical calibrators fit on 180
states (900 action rows).  The 65-D router interface is task plus Medium's
calibrated block-pose consequence.  It is trained on
$5{,}200\times39$ query--task rows; the threshold is selected only from the
$1{,}200\times39$ validation rows.  The primary family contains five
comparisons and 20,000 state-clustered bootstrap repetitions.  An additional
5,000-draw exact-count random-input benchmark preserves every seed's realized
escalation count; it is descriptive and is not part of the inferential family.

V107 is a second, non-overlapping PushT confirmation generated only after its
fresh-state manifest and primary comparison are sealed.  It contains two new
800-state shards, with the same 39 tasks, five candidates, horizon, and three
frozen checkpoint pairs.  The prediction and current-DINO/action routers are
the six direct-gross MLPs selected during development, then copied and hashed
before V107 generation.  The primary family contains exactly one comparison
and 20,000 state-clustered bootstrap repetitions.

\paragraph{Controlled \pybullet{}.}
The frozen confirmation contains six 2,000-state banks: an ID
replication and prespecified changes in horizon, velocity, obstacle count, and
their composition.  The 81-task family crosses nine spatial targets with nine
loss geometries, including anisotropic, diagonal, movement-regularized, and
safety-penalized losses.  Three frozen model seeds yield 36,000 crossed state--seed rows representing
12,000 unique physical states after tasks are averaged.  The selected
59-dimensional task--prediction--regime interface and all thresholds were
frozen from a prespecified development ablation and readiness audit before any
confirmation outcome was opened.  The primary estimand weights banks and
model seeds equally; four comparisons and 10,000 bootstrap repetitions form
the primary family.  The predictors are residual token Transformers over
pooled DINOv2 features: Medium uses width 192, four heads, and two layers;
Full uses width 384, six heads, and three layers.  Training uses AdamW for 100
epochs on 5,000 states, followed by frozen physical calibration on 1,000
states.

\begin{table}[t]
\centering
\caption{Frozen predictive computations and router interfaces.  Parameter/MAC
counts are available for the \pusht{} MLPs; the archived \pybullet{} protocol
records its Transformer dimensions.}
\label{tab:models}
\small
\resizebox{\linewidth}{!}{\begin{tabular}{lrrrr}
\toprule
Evaluation / capacity & Architecture & Parameters & MACs/candidate & Interface dim. \\
\midrule
PushT Medium & MLP 256$\times$2 & 99,392 & 98,816 & 65 \\
PushT Full & MLP 512$\times$3 & 592,448 & 590,848 & --- \\
PyBullet Medium & Transformer 192/4/2 & --- & --- & 59 \\
PyBullet Full & Transformer 384/6/3 & --- & --- & --- \\
\bottomrule
\end{tabular}
}
\end{table}

\paragraph{Latency protocol.}
All costs use one physical state, one downstream task, and five candidate
actions as the decision unit.  World-model forwards were measured on NVIDIA
H100 GPUs.  Task and router components were measured single-threaded on the
associated HPC CPU nodes.  \pusht{} uses 200 warmups, seven rounds of 1,000
repeats, 512 sampled queries, and the median of round medians; p90 accounting
is a sealed sensitivity.  \pybullet{} freezes end-to-end component medians
from its pre-confirmation readiness audit.  Absolute latency portability is
limited because the compact archive does not record the CPU SKU.

\section{Results}
\label{sec:results}

\subsection{Additional capacity has conditional decision value}

\begin{figure}[t]
  \centering
  \includegraphics[width=0.72\linewidth]{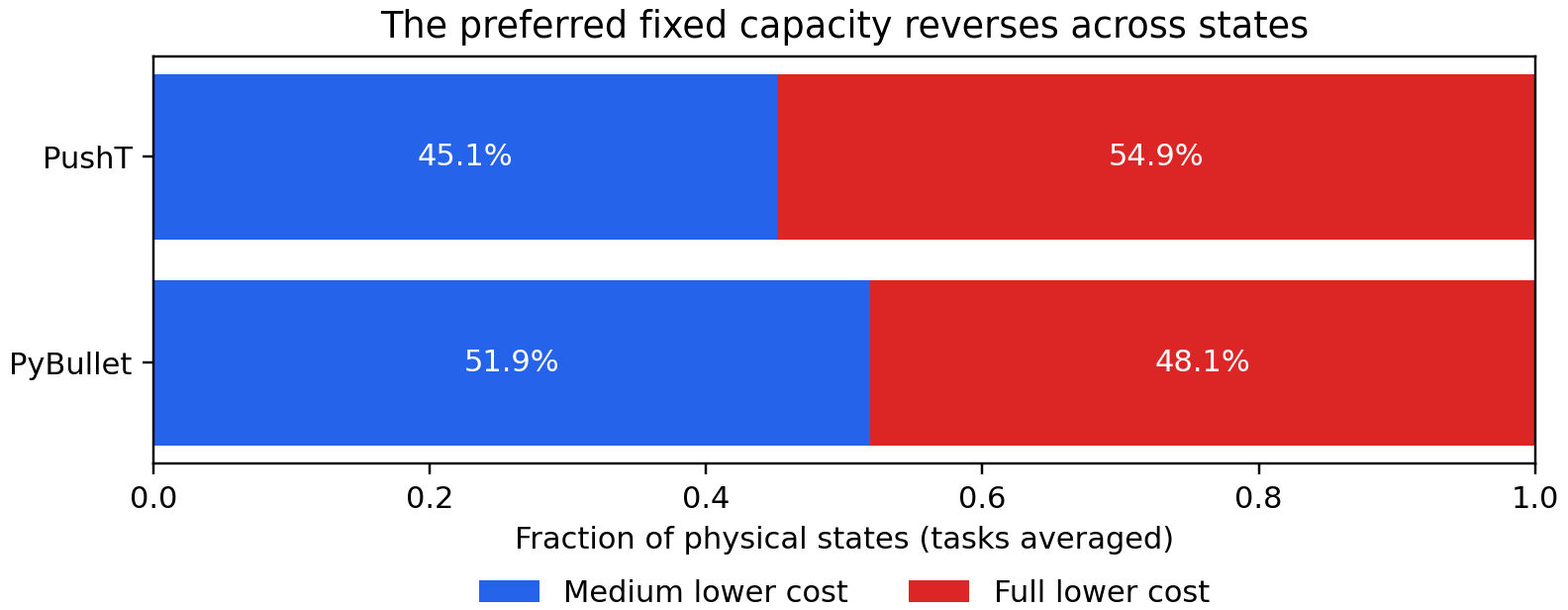}
  \caption{Empirical capacity reversals in the finite audits after averaging
  downstream tasks and the three frozen checkpoint effects within each unique
  physical state.  Medium has lower decision cost on 51.9\% of controlled
  \pybullet{} states and 45.1\% of PushT states; Full wins on the complements.
  These realized reversals do not establish Blackwell incomparability or, by
  themselves, value available to the deployed interface.}
  \label{fig:heterogeneity}
\end{figure}

Figure~\ref{fig:heterogeneity} shows that neither fixed capacity is a
statewise oracle.  After averaging the three frozen seed effects per unique
state, Full wins on 48.1\% of controlled-\pybullet{} states and 54.9\% of
PushT states.  These reversals establish substantial realized state-level
heterogeneity in the finite audit and a clairvoyant state-conditioned selection
opportunity.  They do not by themselves establish positive value for the
declared routing interface; exploitability is assessed by the frozen router
comparisons below.

\subsection{A fresh current-DINO control confirms incremental prediction-interface value}

\begin{figure}[H]
  \centering
  \includegraphics[width=0.82\linewidth]{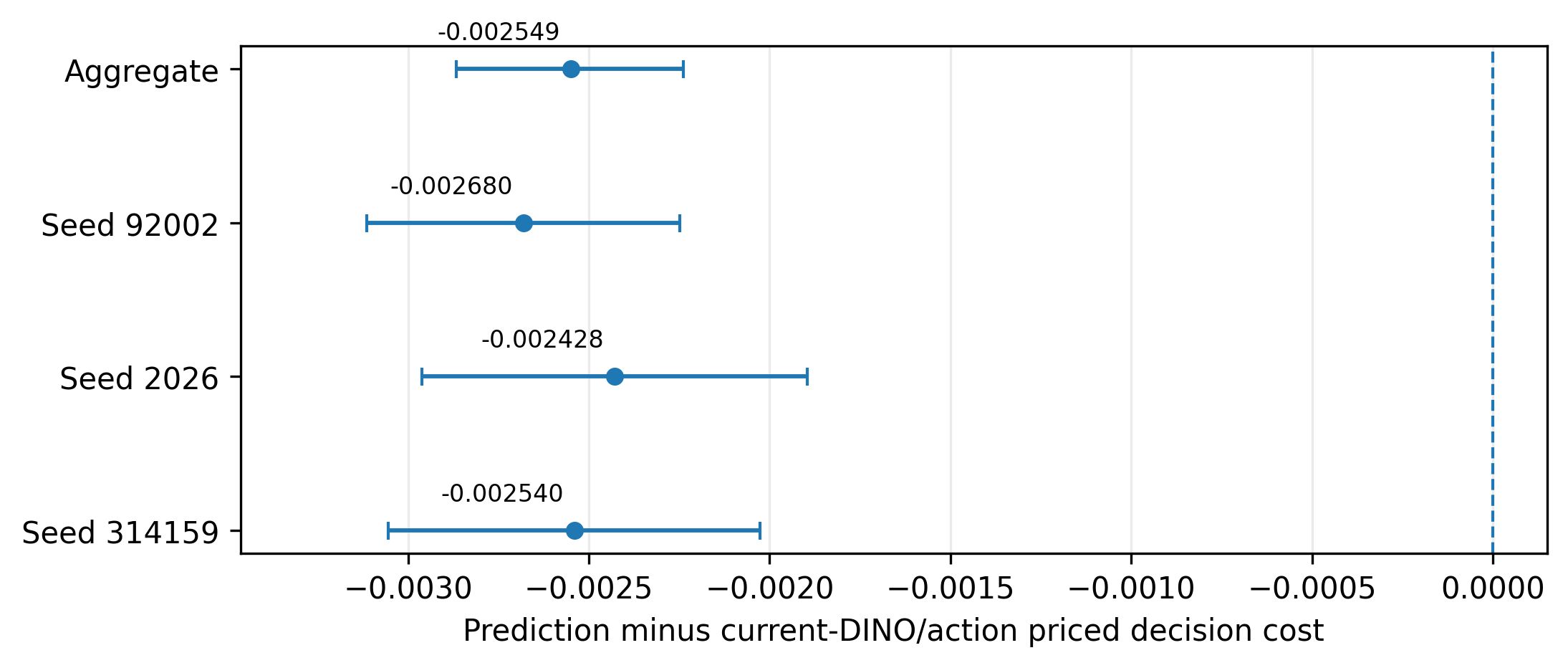}
  \caption{Prospectively sealed V107 prediction-minus-current-DINO/action
  decision-cost effects at $\lambda=0.002$.  The aggregate first averages 39
  tasks and three fixed checkpoint-pair effects within each of 1,600 fresh
  states; bars are state-clustered 95\% bootstrap intervals.  Checkpoint-pair
  rows are fixed-design diagnostics, not a population interval over training
  randomness.  Negative values favour the prediction interface.}
  \label{fig:v107}
\end{figure}

The V107 primary gate passes.  On the second fresh exact-reset PushT bank, the
prediction interface lowers priced physical decision cost relative to the
dimension-matched current-DINO/action control by $0.002549$, with
state-clustered 95\% interval $[-0.002867,-0.002238]$ and one-sided 95\% upper
bound $-0.002286$.  The fixed checkpoint-pair effects are $-0.002680$,
$-0.002428$, and $-0.002540$ for seeds 92002, 2026, and 314159; every
checkpoint-specific 95\% interval is also below zero.  This is the one primary
comparison declared before V107 outcomes, so no exploratory control or learner
can replace the gate.

\begin{table}[H]
\centering
\caption{V107 aggregate operating point at $\lambda=0.002$, equally weighting
the three fixed checkpoint pairs after averaging 39 tasks within each fresh
state.  The control is faster and receives current DINO for free, but the
prediction interface reduces physical regret enough to lower priced cost by
7.9\% relative to the control.}
\label{tab:v107}
\small
\begin{tabular}{lrrrr}
\toprule
Policy & Regret & Latency (ms) & Decision cost & Escalation \\
\midrule
Prediction interface & 0.02849 & 0.664 & 0.02981 & 47.2\% \\
Current-DINO + actions & 0.03110 & 0.630 & 0.03236 & 66.2\% \\
\midrule
Prediction $-$ control & -0.00262 & +0.034 & -0.00255 & -19.0 pp \\
\bottomrule
\end{tabular}

\end{table}

Table~\ref{tab:v107} isolates the mechanism of the priced gain.  Prediction
routing has mean regret $0.02849$ versus $0.03110$ for the control, while its
mean latency is $0.664$\,ms versus $0.630$\,ms.  The $0.002617$ regret
improvement therefore exceeds the $0.000068$ priced latency disadvantage.
Prediction routing also escalates less often, 47.2\% versus 66.2\%.  Because the
control receives the task, current DINO, and all candidate actions with matched
dimension and favourable accounting, this result rules out the narrow
explanation that the V106 gain is obtainable from this current-state/action
interface alone.  It does not prove that imagined futures are causally
sufficient or superior to every possible current-state representation.

\paragraph{Earlier V106 prediction-withheld comparison.}
The initial fresh PushT interface test uses the task-only input router.  At
$\lambda=0.002$, that comparator routes to Full on 100\% of audit query--task
rows for checkpoint pairs 2026 and 314159 and 97.4359\% for checkpoint pair
92002, or 99.1453\% under equal checkpoint weighting.  It is therefore
latency-advantaged but nearly a standalone-Full policy rather than a rich
task-conditioned selector.  Despite paying Medium before routing, the
prediction interface lowers decision cost by $0.002620$, with state-clustered
95\% interval $[-0.003037,-0.002213]$.  It also improves over the matched-rate
margin policy by $0.003294$ and expected matched random by $0.003591$.
Among 5,000 independently generated uniform matched-count input-routing
policies per seed, none attained aggregate cost no larger than the candidate;
the plus-one Monte Carlo benchmark tail fraction is
$1/5001\approx0.00020$.  Because the learned assignment was not randomized and
no row-exchangeability null is assumed, this is descriptive rather than a
hypothesis-test $p$-value.

\paragraph{Controlled \pybullet{}.}
Controlled \pybullet{} supports selective routing for the frozen
prediction--task--regime interface, but does not independently isolate the
prediction from the regime descriptor.  Under repaired fixed-seed
state-clustered inference, candidate-minus-margin is $-0.002800$ (95\% interval
$[-0.002902,-0.002698]$), and candidate-minus-random is $-0.003668$
($[-0.003773,-0.003565]$).  The paired crossed-seed robustness display gives
wider 2.5\%--97.5\% empirical sensitivity quantiles
$[-0.002953,-0.002642]$ and $[-0.004090,-0.003352]$, respectively; these are
not confidence coverage for a broader checkpoint-generating population.

\subsection{The original V106 and PyBullet primary families pass}

\begin{figure}[t]
  \centering
  \includegraphics[width=\linewidth]{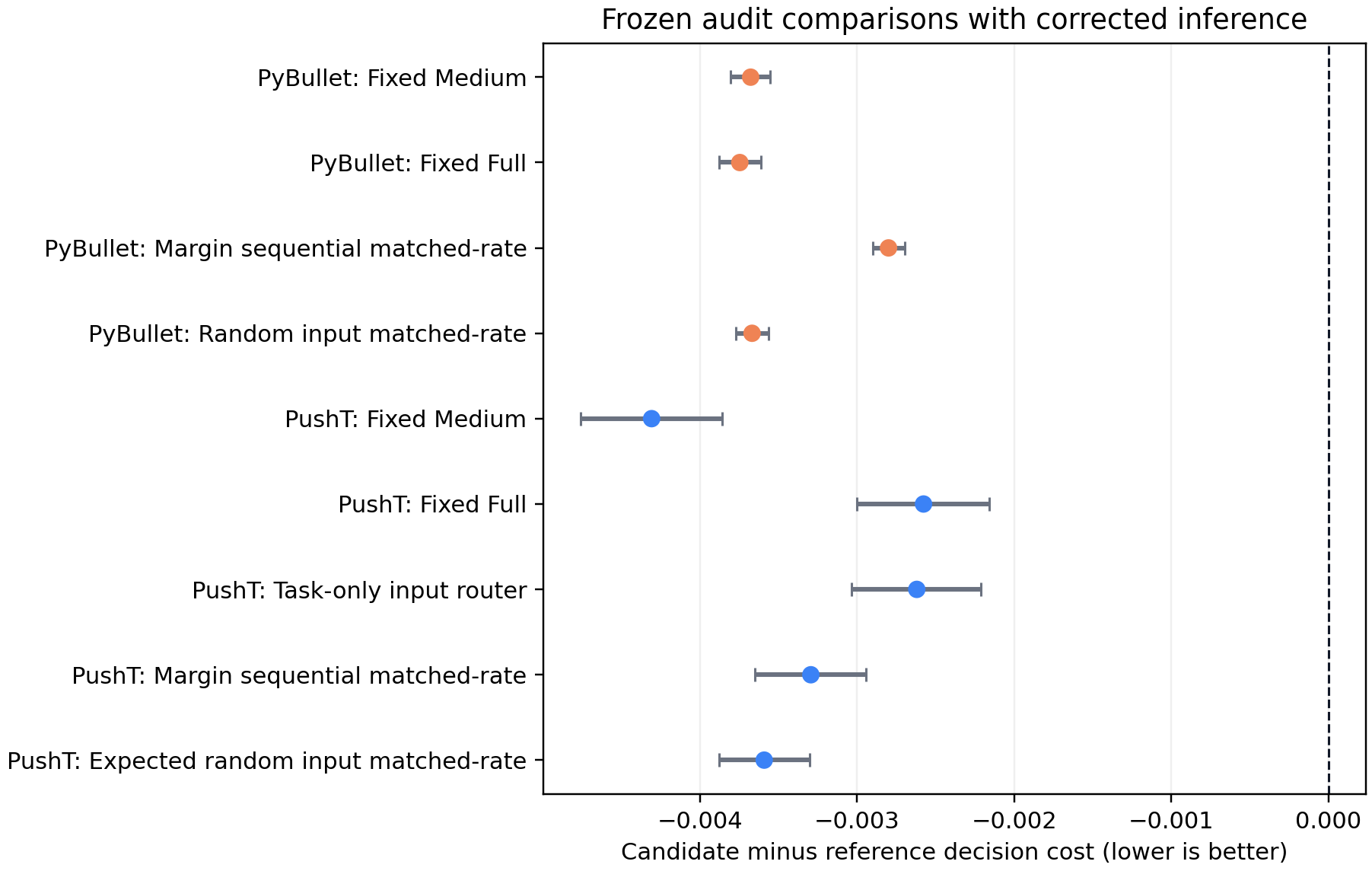}
  \caption{Priced decision-cost improvement at the prespecified
  $\lambda=0.002$ for the nine original V106/PushT and V104E/PyBullet
  comparisons.  The separate V107 primary comparison appears in
  Figure~\ref{fig:v107}.  PushT retains its frozen
  state-clustered intervals; PyBullet uses the repaired fixed-checkpoint
  state-clustered bootstrap.  Negative values favor the router, and all
  environment-specific Bonferroni-adjusted percentile-bootstrap upper bounds
  are below zero.}
  \label{fig:forest}
\end{figure}

Figure~\ref{fig:forest} summarizes all primary comparisons.  On \pusht{}, the
candidate improves over Fixed Medium by $0.004308$
($[-0.004759,-0.003857]$) and Fixed Full by $0.002577$
($[-0.003001,-0.002159]$).  On controlled \pybullet{}, the corresponding unchanged point improvements
are $0.003677$ and $0.003746$, with repaired 95\% intervals
$[-0.003805,-0.003553]$ and $[-0.003880,-0.003613]$.

All three model seeds have favorable point estimates against Fixed Medium
and matched random in each environment.  Every PyBullet bank has a favorable
point estimate against those two references, and no individual unadjusted 95\%
bankwise interval lies entirely above zero against either reference.  Every PushT weight
profile is favorable against both fixed references, with no profile interval
entirely above zero.  Under resampling from the empirical three-checkpoint distribution, the
global PyBullet crossed-seed sensitivity has all four adjusted upper
quantiles below zero.  The corrected PushT crossed sensitivity likewise
remains favorable for all five references, with 2.5\%--97.5\% empirical
quantiles ranging from $[-0.004870,-0.003750]$ against Fixed Medium to
$[-0.003218,-0.001891]$ against Fixed Full.  These are robustness displays,
not confidence intervals for a broader checkpoint-generating population.

\subsection{Decision improvement is not compute saving}

\begin{figure}[t]
  \centering
  \includegraphics[width=\linewidth]{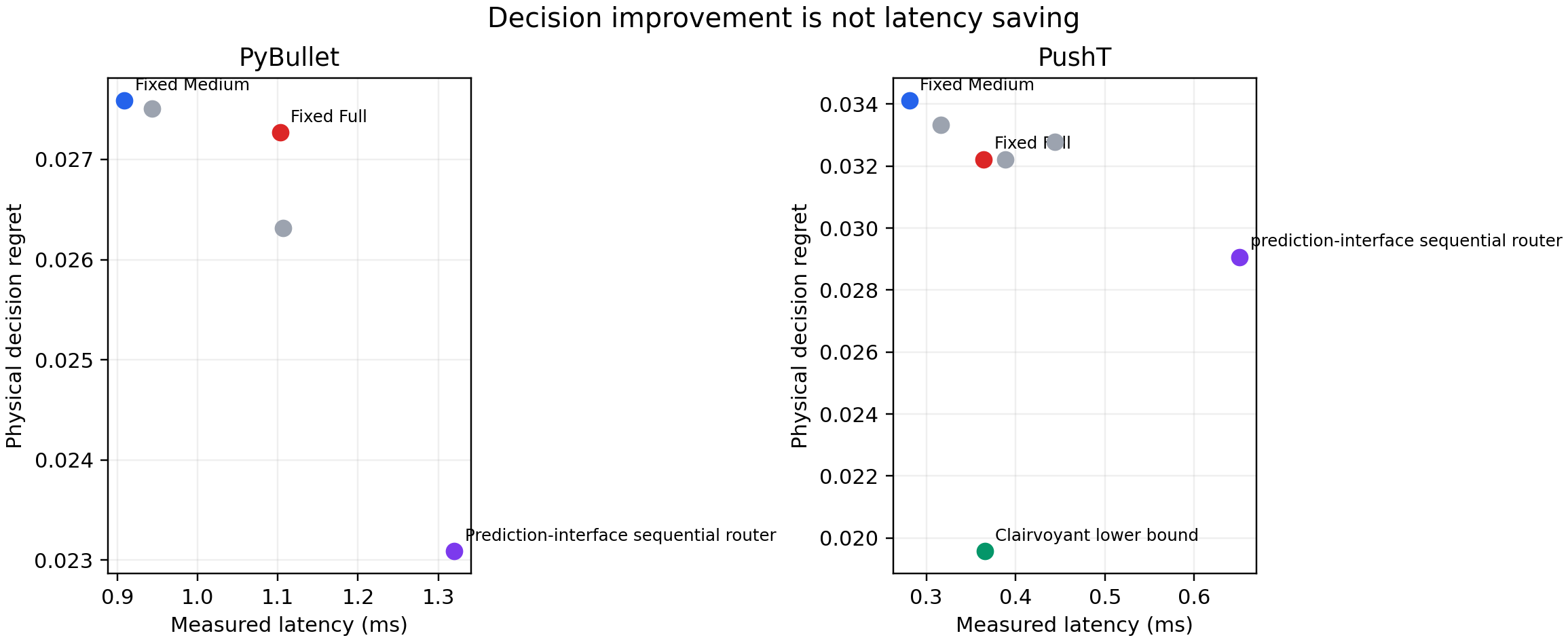}
  \caption{The router is slower than both fixed policies.  It buys a physical
  regret reduction, but its Medium-first interface overhead prevents raw
  latency dominance.  The plotted ``clairvoyant lower bound'' uses realized
  audit outcomes and zero router overhead; it is not the Bayes interface
  oracle.}
  \label{fig:latency}
\end{figure}

On \pusht{}, the router's mean regret is $0.02907$ versus $0.03411$ for Medium
and $0.03222$ for Full, but latency is $0.651$\,ms versus $0.281$ and
$0.364$\,ms.  It escalates 42.0\% of query--task pairs.  Its decision cost
$0.03037$ beats both fixed policies after pricing latency, yet a cheaper clairvoyant empirical lower bound that selects from realized
audit outcomes reaches $0.02031$.  The difference mixes unavailable realized
information, policy estimation error, and protocol-overhead differences; it is
not an estimate of Bayes interface regret.

Controlled \pybullet{} shows the same boundary.  Mean router regret is
$0.02309$ versus $0.02759$ and $0.02727$, while latency is $1.320$\,ms versus
$0.909$ and $1.103$\,ms; the escalation rate is 17.8\%.  These numbers rule
out ``compute saving,'' ``latency dominance,'' and ``near oracle'' as
interpretations of our results.  The measured signed latency gaps relative to
standalone Full are $\E[\kappa_{\G}]=0.49785$ ms in PushT and $1.12352$ ms in
PyBullet, priced as $0.000996$ and $0.002247$ at $\lambda=0.002$.

\begin{table}[t]
\centering
\caption{Primary operating point at $\lambda=0.002$.  The router lowers regret
and priced decision cost while increasing latency.  Environment and seed
weights follow the frozen estimands.}
\label{tab:policies}
\small
\resizebox{\linewidth}{!}{\begin{tabular}{llrrrr}
\toprule
Evaluation & Policy & Regret & Latency (ms) & Decision cost & Escalation \\
\midrule
PyBullet & Router & 0.02309 & 1.320 & 0.02573 & 17.8\% \\
PyBullet & Fixed Medium & 0.02759 & 0.909 & 0.02941 & 0.0\% \\
PyBullet & Fixed Full & 0.02727 & 1.103 & 0.02948 & 100.0\% \\
PushT & Router & 0.02907 & 0.651 & 0.03037 & 42.0\% \\
PushT & Fixed Medium & 0.03411 & 0.281 & 0.03468 & 0.0\% \\
PushT & Fixed Full & 0.03222 & 0.364 & 0.03295 & 100.0\% \\
\bottomrule
\end{tabular}
}
\end{table}

\begin{figure}[t]
  \centering
  \includegraphics[width=0.88\linewidth]{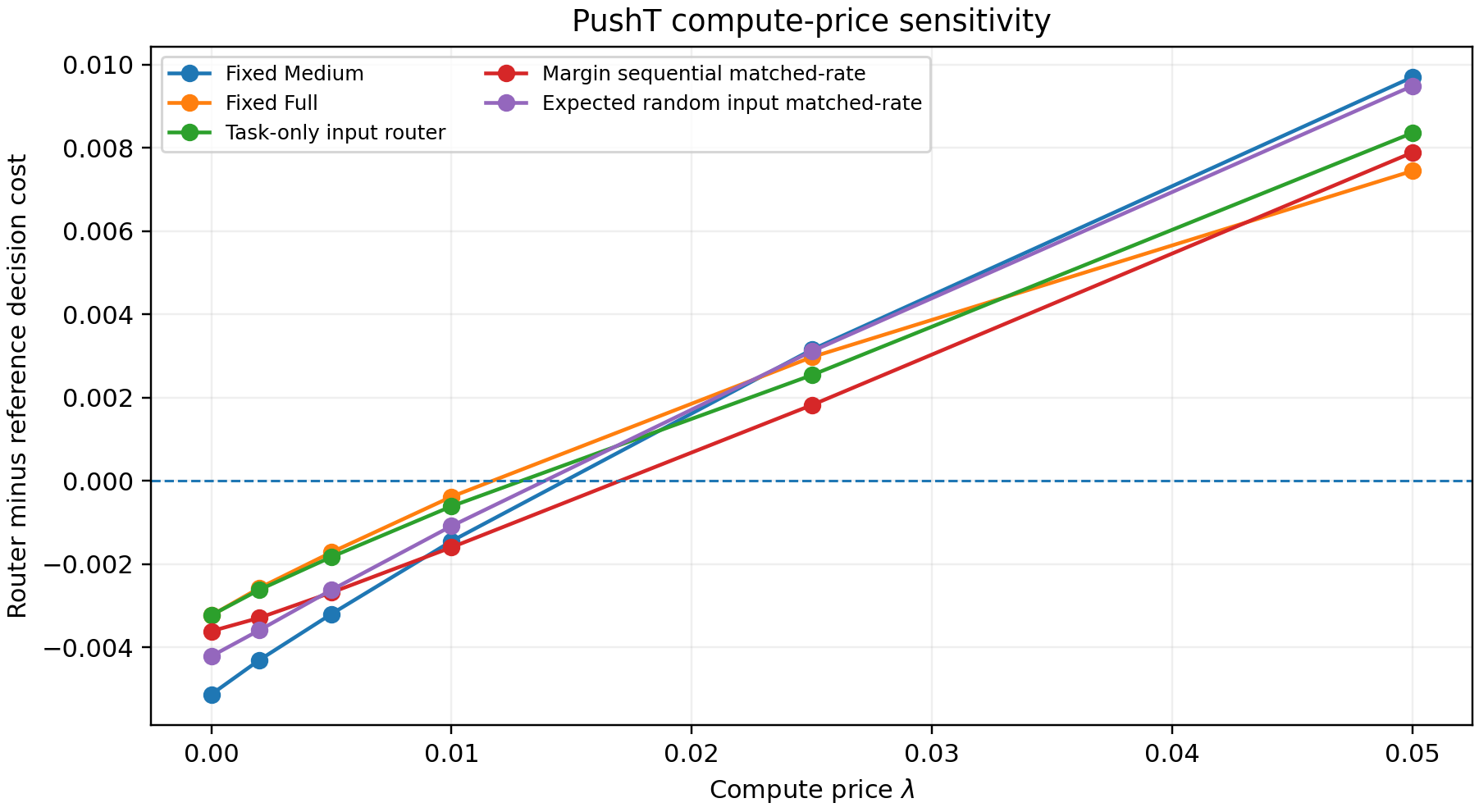}
  \caption{Declared PushT compute-price sensitivity, not a deployment Pareto
  frontier.  Negative router-minus-reference values favor routing.  Thresholds
  are selected on validation separately for each declared price.  The router
  wins at $\lambda\le0.01$ but loses to every listed reference at $0.025$ and
  $0.05$.}
  \label{fig:lambda}
\end{figure}

\begin{table}[t]
\centering
\caption{PushT candidate-minus-reference decision cost across the full declared
compute-price grid.  The primary frozen point is $\lambda=0.002$; positive
values at larger prices delimit rather than invalidate that operating-point
result.}
\label{tab:lambda}
\small
\resizebox{\linewidth}{!}{\begin{tabular}{rrrrrr}
\toprule
$\lambda$ & vs. Medium & vs. Full & vs. task-only & vs. margin & vs. random \\
\midrule
0.000 & -0.0051 & -0.0032 & -0.0032 & -0.0036 & -0.0042 \\
0.002 & -0.0043 & -0.0026 & -0.0026 & -0.0033 & -0.0036 \\
0.005 & -0.0032 & -0.0017 & -0.0018 & -0.0027 & -0.0026 \\
0.010 & -0.0015 & -0.0004 & -0.0006 & -0.0016 & -0.0011 \\
0.025 & +0.0031 & +0.0030 & +0.0025 & +0.0018 & +0.0031 \\
0.050 & +0.0097 & +0.0074 & +0.0084 & +0.0079 & +0.0095 \\
\bottomrule
\end{tabular}
}
\end{table}
\clearpage

Figure~\ref{fig:lambda} and Table~\ref{tab:lambda} show that the result is not a
continuous Pareto improvement.  The candidate remains favorable through
$\lambda=0.01$, with a narrow $-0.000385$ difference versus Fixed Full at that
price, but becomes worse than all fixed and routing references at
$\lambda\in\{0.025,0.05\}$.  In the separate V107 fixed-threshold
sensitivity, prediction remains below the current-DINO/action router throughout
$\lambda\in\{0,0.002,0.005,0.01,0.025,0.05\}$, but this does not imply
high-price deployment value: both routing policies can be dominated by fixed
policies as latency becomes expensive.  Scientific replay uses the exact serialized
threshold strings cast to NumPy float32 and a strict
$\text{score}>\text{threshold}$ comparison.  Across the task-only grid there
are 17,600 score--threshold equalities.  A non-scientific replay that reads
those thresholds through the default pandas float parser and promotes scores
to float64 changes 6,400 high-price decisions; the released all-grid audit,
mask hashes, and regression test lock the scientific convention.

\section{Related work}

\paragraph{World models and predictive planning.}
DINO-WM predicts future DINOv2 patch features and plans zero-shot across
navigation, manipulation, and multi-particle tasks
\citep{zhou2024dinowm,ocquab2023dinov2}.  Our experiments likewise use frozen
semantic visual features, but ask how the predicted consequence should inform
selection between frozen computations rather than how to learn a planner.
Decision-focused learning instead trains predictions for downstream
optimization loss \citep{wilder2019melding}; prediction-interface routing
leaves the world models frozen and learns the conditional value of switching
between them.  ACID modifies the planning score with inverse-dynamics action
consistency across several latent and video world models
\citep{seo2026acid}; we instead leave each downstream scorer frozen and ask
which predictive computation to evaluate.

\paragraph{Adaptive WAM inference.}
Several concurrent systems already establish prediction- or
intermediate-state-conditioned computation.  Gated geometric Best-of-$N$ uses
an initial action and predicted visual future to decide whether to draw
additional world-action-model rollouts, includes a same-trigger-rate random
diagnostic, and evaluates closed-loop task success across multiple backbones
\citep{zhao2026geometric}.  SANTS learns state-dependent stopping along a video
denoising trajectory and reports closed-loop latency reductions
\citep{sun2026sants}; adaptive action execution invokes another model pass when
prediction and reality diverge \citep{wang2026imagination}; and adaptive-depth
latent world models expose help, hurt, and flat depth regimes during planning
\citep{sivasankar2026adaptive}.  Fast-WAM questions whether explicit future
generation is needed at inference, while AHA-WAM routes and reuses long-horizon
world context asynchronously \citep{yuan2026fastwam,cai2026ahawam}.  These
works establish the broad adaptive-compute premise.  Our difference is not that
premise but the combination of separate frozen capacities, paired exact-reset
downstream losses, task-dependent decision evaluation, and duplication-aware
comparison with standalone Full.

\paragraph{Adaptive cascades, deferral, and information value.}
Adaptive neural networks already learn continuation from a cheap stage signal
using per-example cheap-versus-expensive losses and runtime
\citep{bolukbasi2017adaptive}.  Learning to defer and post-hoc deferral select
between fixed experts from conditional risk differences
\citep{mozannar2020consistent,narasimhan2022posthoc}; confidence-based cascade
analysis characterizes when a cheap model's confidence is insufficient
\citep{jitkrittum2023confidence}; and consistent two-stage acquisition studies
whether paid additional information is worth its cost
\citep{regol2025acquisition}.  Rational metareasoning studies computational
actions \citep{russell1991principles}, while conditional-risk calibration
treats the required regression problem directly
\citep{vasilyev2026conditional}.  Our Bayes sign rule, positive-part value,
weighted sign regret, and fixed-rate upper-tail result are standard or adapted
specializations.  We use them as an estimand and cost ledger; the distinctive
evidence is the paired physical protocol and frozen audit comparison.

\paragraph{Calibration and statistical experiments.}
Decision calibration evaluates predictions through their induced decision
losses \citep{zhao2021calibrating,zhao2021right}.  Blackwell comparison, Le Cam
deficiency, and deficiency-based representation analysis provide general
languages for informativeness
\citep{blackwell1953equivalent,lecam2000asymptotics,
vanrooyen2014lecam,banerjee2018variational}.  Our current experiments
do not estimate a garbling kernel, deficiency, or a separating class of all
decision problems; rescue and reversal events therefore motivate but do not
establish statistical-experiment incomparability.

\section{Limitations and conclusion}

The action sets and downstream loss families are controlled, and none of the
evaluations is a closed-loop robotics deployment.  The predictors remain
related DINO-style calibrated interfaces rather than genuinely different
world-model families, and only three fixed checkpoint pairs are available.
V107 answers one important alternative explanation prospectively: under its
tested 65-dimensional interfaces, the aligned Medium prediction contains
routing information not recovered by a task-, current-DINO-, and
action-conditioned input router that is deliberately favoured in latency.
This does not establish prediction causality, sufficiency, or superiority to
all current-state encodings, architectures, or compute-matched ensembles.

Full is not uniformly better, benefit prediction is imperfect, and the
clairvoyant lower-bound gap remains substantial.  Timing combines measured
components on a specified H100-based stack but is not synchronized end-to-end
deployment evidence; the sequential prediction router is slower than fixed
policies.  The original PushT fixed-policy sensitivity supports only a
low-price operating region, not a continuous frontier.  PyBullet validates a
composite task--prediction--regime interface and does not independently isolate
prediction from regime.

Within those boundaries, the evidence supports a sharper interface claim than
the initial audit alone.  Under paired exact-reset outcomes, the tested
Medium-derived prediction interface supports useful post-hoc selection between
two frozen predictors.  It beats fixed and prediction-withheld references on
the first fresh PushT bank, and on a second prospectively sealed bank it beats
a dimension-matched, action-conditioned current-DINO router with no encoder
latency charged.  All three fixed checkpoint-pair effects are favourable, and
the primary state-clustered gate passes.  Controlled PyBullet supplies a second
physical mechanism, with its conclusions surviving a transparent repair of the
crossed-state bootstrap.

The contribution remains a paired physical evaluation and audit protocol, not
a new generic routing theory.  The study does not establish compute saving,
closed-loop value, or generality across world-model families.  The highest-value
main-track extension is now a genuinely different stochastic or iterative
world-model family with nested additional computation, closed-loop evaluation,
and synchronized end-to-end timing.

\clearpage
\appendix

\section{Proofs}

\subsection{Proof of the standard Bayes escalation result (Proposition~\ref{prop:bayes})}
Conditioning Eq.~\eqref{eq:routercost} on $\G$ gives
\[
 \E[C_{\G}(\pi)\mid\G]
 =
 \E[L_M+\lambda s_{\G}\mid\G]-\pi b_{\G}.
\]
Since $\pi\in\{0,1\}$, the pointwise minimizer escalates exactly when
$b_{\G}>0$.  Substitution yields
\[
 \E[C_{\G}(\pi^\star_{\G})\mid\G]
 =
 \E[L_M+\lambda s_{\G}\mid\G]-\pos{b_{\G}}.
\]
Taking expectations proves the claim.

\subsection{Proof of the accounting identity (Proposition~\ref{prop:standalone})}
For Fixed Medium, subtract Proposition~\ref{prop:bayes}'s cost:
\[
\E[C_M^{\mathrm{fix}}-C_{\G}(\pi^\star_{\G})]
=
\E[L_M+\lambda c_M-L_M-\lambda s_{\G}]+V(\G),
\]
which is Eq.~\eqref{eq:gainmedium}.  Always escalating under the sequential
protocol costs $L_F+\lambda(s_{\G}+d_F)$.  Its excess over the Bayes router is
\[
-\E[B]+V(\G)
=-\E[b_{\G}]+\E[\pos{b_{\G}}]
=\E[\pos{-b_{\G}}].
\]
Standalone Full removes $\lambda\kappa_{\G}$ from that always-escalate cost,
which gives Eq.~\eqref{eq:gainfull}.

\subsection{Proof of the nested-interface Jensen identity (Theorem~\ref{thm:piv})}
Nested conditioning gives
\[
 \E[b_{\G_1}\mid\G_0]=b_{\G_0}.
\]
The positive-part map $z\mapsto z_+$ is convex, so conditional Jensen implies
\[
 \pos{b_{\G_0}}
 \leq \E[\pos{b_{\G_1}}\mid\G_0].
\]
Taking expectations proves nonnegativity.  Let $U=b_{\G_1}$ and define
$P_0=\E[U_+\mid\G_0]$ and $N_0=\E[(-U)_+\mid\G_0]$.  Since
$\E[U\mid\G_0]=P_0-N_0$, the conditional Jensen gap is
$P_0-(P_0-N_0)_+=\min(P_0,N_0)$.  Its expectation is positive exactly when a
positive-probability set has both $P_0>0$ and $N_0>0$.  Subtracting the
interface-overhead terms gives the deployment comparison.

\subsection{Proof of the standard sign-regret identity (Proposition~\ref{prop:plugin})}
Condition on $\G$.  A correct sign choice matches the Bayes action and has
zero excess cost.  A sign mismatch forfeits exactly $|b_{\G}|$.  Therefore the
conditional excess is
$|b_{\G}|\mathbf{1}\{\widehat\pi\neq\pi^\star_{\G}\}$.
On a sign mismatch, zero lies between $\widehat b$ and $b_{\G}$, implying
$|b_{\G}|\leq|\widehat b-b_{\G}|$.  Take expectations.

\subsection{Proof of Corollary~\ref{cor:learned}}
Proposition~\ref{prop:plugin} shows that replacing
$\pi^\star_{\G}$ by $\widehat\pi$ increases expected sequential cost by
exactly $\mathcal R_{\G}(\widehat\pi)$.  Subtracting this amount from each
Bayes gain in Proposition~\ref{prop:standalone} gives
Eqs.~\eqref{eq:learnedmedium}--\eqref{eq:learnedfull}.

\subsection{Proof of Corollary~\ref{cor:margin}}
If $\varepsilon=0$, sign disagreement can occur only where $b_{\G}=0$, so the
weighted excess cost is zero.  Assume henceforth that $\varepsilon>0$.
A sign error implies $|b_{\G}|\le\varepsilon$.  Hence the exact identity is at
most
\[
\E\!\left[|b_{\G}|\mathbf 1\{|b_{\G}|\le\varepsilon\}\right]
\le \varepsilon\Prb(|b_{\G}|\le\varepsilon)
\le A\varepsilon^{\alpha+1}.
\]

\subsection{Proof of the scalar-ordering obstruction (Proposition~\ref{prop:threshold})}
Let
$z_+=\operatorname*{ess\,sup}_{A_+}Z$ and
$z_-=\operatorname*{ess\,inf}_{A_-}Z$, so $z_+<z_-$.
If $\tau<z_-$, the threshold escalates almost surely on $A_-$ and therefore
disagrees with the Bayes rule there.  If $\tau\geq z_-$, then
$\tau>z_+$, so it does not escalate almost surely on $A_+$ and disagrees
there.  The exact identity in Proposition~\ref{prop:plugin} therefore lower
bounds the excess cost by $\gamma\Prb(A_-)$ in the first case and by
$\gamma\Prb(A_+)$ in the second.  Taking the smaller bound proves the claim.

\subsection{Proof of the fixed-rate rearrangement lemma (Proposition~\ref{prop:matched})}
The endpoint cases use $g^\star\equiv0$ or $g^\star\equiv1$.  For
$0<q<1$, choose $t$ such that
$\Prb(b_{\G}>t)\le q\le\Prb(b_{\G}\ge t)$ and choose
$\rho\in[0,1]$ so that
$g^\star=\mathbf1\{b_{\G}>t\}+\rho\mathbf1\{b_{\G}=t\}$ has
$\E[g^\star]=q$.  Conditional expectation gives $\E[gB]=\E[gb_{\G}]$.
For any feasible $g$ with $\E[g]=q$, we have $\E[g^\star-g]=0$, and therefore
\[
 \E[b_{\G}(g^\star-g)]
 =\E[(b_{\G}-t)(g^\star-g)]\ge0.
\]
The integrand is nonnegative on $\{b_{\G}>t\}$ and
$\{b_{\G}<t\}$, and zero on $\{b_{\G}=t\}$.  Hence $g^\star$
maximizes $\E[b_{\G}g]$.  Its advantage over independent rate-$q$ routing is
$\mathrm{Cov}(b_{\G},g^\star)\ge0$, with strict inequality for
$0<q<1$ exactly when $b_{\G}$ is not almost surely constant.  Independent uniform randomization
realizes the propensity as a binary action on
$\G\vee\sigma(U)$ without changing its conditional expectation.

\subsection{Proof of the finite-record observation (Proposition~\ref{prop:paired})}
Every frozen model decision and policy action is replayable from a complete
executed record.  Because every $Y(Q,a)$ is present there, both $L_M(Q)$ and
$L_F(Q)$, hence $B(Q)$ and each realized policy cost, are deterministic
functions of that record.  Finite-audit means are therefore exactly
computable.  The public compact package does not claim to be this complete
candidate-level record; it releases the policy-facing arrays needed to
recompute the reported statistics.  A declared target law and weighting rule
separately define a population estimand; when the required costs are observable
functionals of the cluster law, that law identifies the expectation.  Sampling assumptions then justify consistency and
asymptotics, while bootstrap coverage requires its own regularity conditions
and a resampling scheme that preserves all clustered and crossed factors.

\section{Why latent distance is not a common currency}

\begin{proposition}[Measurable recoding observation]
\label{prop:coordinates}
Let $Z$ take values in a finite discrete subset
$\{z_1,\ldots,z_n\}$ of a standard-Borel space.  For any distinct Euclidean
target vectors $w_1,\ldots,w_n$, define $T(z_i)=w_i$.  Then $T$ and its inverse
on the image are measurable, $\sigma(T(Z))=\sigma(Z)$, and every Bayes risk
over unrestricted measurable decision rules is unchanged.  Coordinate-dependent
geometric statistics need not be preserved.  The conclusion does not preserve
a restricted neural class, sample complexity, regularization, optimization,
continuity, or computational cost.
\end{proposition}

\paragraph{Proof.}
The assignment $T(z_i)=w_i$ is one-to-one because the target vectors are
distinct.  On finite supports both $T$ and its inverse are measurable, and
each is a deterministic garbling of the other.  They therefore generate the
same sigma-field.  Any decision rule based on one signal can be composed with
the inverse to obtain a rule with identical loss under the other, so all Bayes
risks agree. \qed

Applying model-specific recodings shows why a universal comparison across
independently parameterized latent spaces cannot rest only on Euclidean norm,
pairwise distance, cosine, or a similarly coordinate-dependent geometry.  The
statement is exact for finite supports and extends to bimeasurably equivalent
signals.  It is a representation caution, not a claim that our two models are
Blackwell- or Le-Cam-incomparable: the proposition compares two recodings of
one signal, whereas the experiments compare distinct predictive computations.

\section{Protocol details}

\subsection{Information restrictions}
The \pusht{} prediction candidate feature groups are the task and
Medium's calibrated prediction.  Explicit state, future state, current or
future raw DINO latent, true loss, oracle action, Medium/Full realized regret,
oracle benefit, Full prediction, correctness, coverage, and goal pose are
forbidden to that candidate.

V107 adds one prespecified input-routing control.  It receives the same task,
the current 64-D DINO representation, and all five candidate actions, then
compresses the non-task block to 43 dimensions using a scaler and PCA fit only
on V106 training states.  It receives no predicted future, Full prediction,
true future, or realized loss.  Treating current DINO as already available and
routing before Medium deliberately favours the control.

The controlled-\pybullet{} candidate feature groups are task, predicted
physical consequence, and prespecified regime.  Its upstream ablation removed
explicit simulator state.  The PushT prediction candidate and PyBullet
candidate route only after mandatory Medium computation; the V107
current-DINO/action control is the declared input-routing exception.

\subsection{Sealing and statistical gates}
V107 is sealed after the V106 audit has been used only as development data.
Its protocol, two-shard fresh-state manifest, six router hashes, PCA projection,
thresholds, latency components, float32 scoring implementation, single primary
comparison, and bootstrap seed are fixed before any V107 physical outcome.
Generation and inference verify the seal before execution; audit writes to a
staging directory, verifies all lineage hashes, moves atomically, and repeats
the postcheck.  The primary gate requires a negative aggregate point, a
negative one-sided 95\% upper bound, and negative points for all three fixed
checkpoint pairs.

V106 uses development shards 0--7 and audit-only shards 8--9.  The router,
thresholds, latency components, comparison family, bootstrap scheme, and
matched-count benchmark generator were sealed before audit generation.  Its global gate
requires all five Bonferroni-adjusted percentile-bootstrap one-sided upper
bounds below zero.  Seed replication and weight-profile robustness are
separate frozen criteria.  The matched-count random-input benchmark is also
reported as a prespecified descriptive criterion, but is not an inferential
gate or part of the five-comparison family.

V104E is a fresh confirmation after the V104D interface ablation and V104D2
readiness audit.  The frozen candidate may not use V104E labels for training,
threshold, feature, or hyperparameter selection.  Its global four-reference
gate, seed replication, and six-bank robustness are jointly required.

\subsection{Executed estimands and inference}
For a candidate/reference pair, let $D_{rit}$ denote candidate-minus-reference
cost for model seed $r$, physical state $i$, and task $t$.  V107 uses this
statistic for prediction minus current-DINO/action with
$(n,R,T)=(1600,3,39)$.  It first averages tasks and the three fixed checkpoint
effects within state, then resamples the 1,600 state effects for 20,000
bootstrap repetitions with seed 107031.  Its one-sided upper quantile is 0.95
because there is one primary comparison.  The two-sided interval is reported
descriptively, and checkpoint-specific intervals are fixed-design
diagnostics.

The original V106 \pusht{} primary statistic is
\[
 \widehat\Delta_{\mathrm{PT}}
 =
 \frac1n\sum_{i=1}^{n}
 \frac1R\sum_{r=1}^{R}
 \frac1T\sum_{t=1}^{T}D_{rit},
 \qquad (n,R,T)=(1600,3,39).
\]
The primary bootstrap first forms the checkpoint- and task-averaged state
effect and resamples its $n$ physical states with replacement.  Thus tasks
sharing a state are never treated as independent, and the three checkpoint
effects are fixed design replicates in this primary interval.  The ordinary
interval uses the $0.025$ and $0.975$ bootstrap quantiles; the one-sided
five-comparison Bonferroni-adjusted percentile-bootstrap upper bound uses
quantile $1-0.05/5=0.99$, with approximate simultaneous interpretation under
the stated cluster-sampling model.  A separate crossed sensitivity resamples
checkpoint positions and one common physical-state index vector shared across
all sampled positions; its 2.5\%--97.5\% values are empirical sensitivity
quantiles for the observed three-checkpoint distribution, not population
confidence intervals.

For \pybullet{}, write $D_{brit}$ for bank $b$.  Its unchanged point
estimand is
\[
 \widehat\Delta_{\mathrm{PB}}
 =
 \frac1{BR}\sum_{b=1}^{B}\sum_{r=1}^{R}
 \frac1{n_b}\sum_{i=1}^{n_b}
 \frac1T\sum_{t=1}^{T}D_{brit},
 \qquad (B,R,n_b,T)=(6,3,2000,81).
\]
The repaired primary first averages tasks and the three fixed checkpoint
effects for each physical state, resamples the $n_b=2000$ state clusters
within each fixed bank, and then averages the six bank means.  A crossed
sensitivity resamples three checkpoint positions and, within each bank, one
common state-index vector shared across all sampled positions.  Both use
10,000 replicates; the four-comparison primary analysis uses a
Bonferroni-adjusted one-sided percentile-bootstrap upper bound at quantile
$1-0.05/4=0.9875$, with approximate simultaneous interpretation under the
cluster-sampling model.  Crossed outputs are labelled empirical sensitivity
quantiles.  The originally executed cellwise bootstrap is released for
provenance but is not used for paper-facing uncertainty.

The \pusht{} matched-count random-input benchmark samples, independently for
each model seed, exactly as many query--task rows as the candidate escalated.  With
$R_{\mathrm{perm}}=5000$ draws, the plus-one benchmark tail fraction is
\[
 \widehat p_{\mathrm{bench}}=
 \frac{1+
 \#\{\text{random matched-count cost}\leq\text{candidate cost}\}}
 {R_{\mathrm{perm}}+1}.
\]
This addresses selection beyond escalation count relative to the specified
random-policy generator.  It is descriptive rather than a design-based
hypothesis-test $p$-value, and it is not the source of the state-clustered
confidence intervals.

\section{Complete primary comparisons}

\subsection{Prospective V107 confirmation}
\begin{center}
  \small
  \resizebox{0.92\linewidth}{!}{\begin{tabular}{lrrr}
\toprule
Fixed design & $\widehat{\Delta C}$ & 95\% interval & One-sided 95\% upper \\
\midrule
Aggregate & -0.002549 & [-0.002867, -0.002238] & -0.002286 \\
92002 & -0.002680 & [-0.003113, -0.002248] & -0.002313 \\
2026 & -0.002428 & [-0.002962, -0.001896] & -0.001986 \\
314159 & -0.002540 & [-0.003054, -0.002027] & -0.002110 \\
\bottomrule
\end{tabular}
}
\end{center}

\subsection{Original V106 and V104E families}
\noindent Intervals are state-clustered for both environments; PyBullet uses the
repaired fixed-seed, bank-stratified physical-state bootstrap.
\begin{center}
  \small
  \resizebox{\linewidth}{!}{\begin{tabular}{llrrr}
\toprule
Environment & Reference & $\widehat\Delta C$ & 95\% CI & Gate \\
\midrule
PushT V107 & Current-DINO + actions & -0.0025 & [-0.0029, -0.0022] & Pass \\
PyBullet & Fixed Medium & -0.0037 & [-0.0038, -0.0036] & Pass \\
PyBullet & Fixed Full & -0.0037 & [-0.0039, -0.0036] & Pass \\
PyBullet & Margin sequential matched-rate & -0.0028 & [-0.0029, -0.0027] & Pass \\
PyBullet & Random input matched-rate & -0.0037 & [-0.0038, -0.0036] & Pass \\
PushT & Fixed Medium & -0.0043 & [-0.0048, -0.0039] & Pass \\
PushT & Fixed Full & -0.0026 & [-0.0030, -0.0022] & Pass \\
PushT & Task-only input router & -0.0026 & [-0.0030, -0.0022] & Pass \\
PushT & Margin sequential matched-rate & -0.0033 & [-0.0037, -0.0029] & Pass \\
PushT & Expected random input matched-rate & -0.0036 & [-0.0039, -0.0033] & Pass \\
\bottomrule
\end{tabular}
}
\end{center}

\section{V107 fixed-threshold price sensitivity}

\noindent The V107 routers and thresholds remain frozen at their
$\lambda=0.002$ values.  Negative values favour prediction over the
current-DINO/action control; this comparison does not assert dominance over
fixed policies.
\begin{center}
  \small
  \begin{tabular}{rrr}
\toprule
$\lambda$ & Prediction $-$ control & 95\% interval \\
\midrule
0.000 & -0.002617 & [-0.002934, -0.002306] \\
0.002 & -0.002549 & [-0.002867, -0.002238] \\
0.005 & -0.002448 & [-0.002765, -0.002136] \\
0.010 & -0.002279 & [-0.002598, -0.001967] \\
0.025 & -0.001772 & [-0.002094, -0.001456] \\
0.050 & -0.000927 & [-0.001258, -0.000601] \\
\bottomrule
\end{tabular}

\end{center}

\section{Claim guardrails}

\begin{center}
\small
\begin{tabular}{
  >{\raggedright\arraybackslash}p{0.31\linewidth}
  >{\raggedright\arraybackslash}p{0.62\linewidth}}
\toprule
Unsupported wording & Supported replacement \\
\midrule
The router saves compute. &
The router lowers a priced decision objective but is slower than both fixed
policies. \\
Full is the better expert. &
The value of Full is state- and task-dependent; both fixed capacities win on
substantial subsets. \\
The models are Blackwell-incomparable. &
Capacity ordering reverses empirically for the evaluated decision families. \\
The router is near oracle. &
A substantial gap remains to the realized-loss clairvoyant lower bound, which
uses unavailable audit outcomes and zero router overhead. \\
Prediction is universally superior to current state. &
On one prospectively sealed PushT bank, the tested aligned prediction
interface beats one dimension-matched current-DINO/action router. \\
The method generalizes to arbitrary world models. &
Two controlled physical mechanisms and two fresh PushT banks support the claim
under related calibrated DINO-style interfaces. \\
We establish a continuous frontier. &
We report a frozen primary compute price and separately declared sensitivity
points. \\
\bottomrule
\end{tabular}
\end{center}

\section{Reproducibility and archival note}

The public release candidate separates three layers:
\begin{enumerate}
  \item exact hash-verified frozen sources used by the sealed runs;
  \item compact sealed protocols, manifests, comparisons, and policy summaries;
  \item a new reusable implementation of decision value, interface value,
  routing cost, clustered bootstrap, and lineage checks.
\end{enumerate}
The reusable implementation was not retroactively substituted for the
executed code.  The release now includes the executed V104D and V104D2 scripts,
reports, router checkpoints, thresholds, manifests, the V104E and V106 sealed
result tars in their respective provenance directories, and the compact
per-state arrays required to reconstruct all paper-facing statistics.  A release-mode postcheck maps historical absolute source paths to
their hash-matched package copies while preserving the original frozen
postcheck for provenance.

The archive still does not include the heavy raw observations, all-candidate
physical outcome tensors, DINO feature banks, upstream world-model checkpoints,
or the original HPC environment needed to regenerate every rollout from first
principles.  Consequently, the statistics, frozen router lineage, and source
semantics are independently auditable from the package, whereas full physical
outcome regeneration remains an external-asset task.  Hashes and self-authored
seal records establish released-record consistency, not external timestamped
certification of chronology.  Candidate-action ties use NumPy's deterministic
first-minimum \texttt{argmin} convention; the legacy V106
\texttt{tie\_tolerance} configuration field is not used in action selection.
These are archival disclosures, not scientific reruns.

\begingroup
\small
\sloppy
\setlength{\bibsep}{1.5pt}
\bibliographystyle{plainnat}
\bibliography{references}

\begin{thebibliography}{23}
\providecommand{\natexlab}[1]{#1}
\providecommand{\url}[1]{\texttt{#1}}
\expandafter\ifx\csname urlstyle\endcsname\relax
  \providecommand{\doi}[1]{doi: #1}\else
  \providecommand{\doi}{doi: \begingroup \urlstyle{rm}\Url}\fi

\bibitem[Banerjee and Mont{\'u}far(2018)]{banerjee2018variational}
Pradeep~Kr. Banerjee and Guido Mont{\'u}far.
\newblock The variational deficiency bottleneck.
\newblock \emph{arXiv preprint arXiv:1810.11677}, 2018.
\newblock URL \url{https://arxiv.org/abs/1810.11677}.

\bibitem[Blackwell(1953)]{blackwell1953equivalent}
David Blackwell.
\newblock Equivalent comparisons of experiments.
\newblock \emph{The Annals of Mathematical Statistics}, 24\penalty0
  (2):\penalty0 265--272, 1953.

\bibitem[Bolukbasi et~al.(2017)Bolukbasi, Wang, Dekel, and
  Saligrama]{bolukbasi2017adaptive}
Tolga Bolukbasi, Joseph Wang, Ofer Dekel, and Venkatesh Saligrama.
\newblock Adaptive neural networks for efficient inference.
\newblock In \emph{Proceedings of the 34th International Conference on Machine
  Learning}, volume~70, pages 527--536, 2017.
\newblock URL \url{https://proceedings.mlr.press/v70/bolukbasi17a.html}.

\bibitem[Cai et~al.(2026)Cai, Ling, Chu, Liu, Kang, Liang, Xu, Mao, Zhang,
  Yang, Ying, Zheng, and Mu]{cai2026ahawam}
Jisong Cai, Long Ling, Shiwei Chu, Zhongshan Liu, Jiayue Kang, Zhixuan Liang,
  Wenjie Xu, Yinan Mao, Weinan Zhang, Xiaokang Yang, Ru~Ying, Ran Zheng, and
  Yao Mu.
\newblock {AHA-WAM}: Asynchronous horizon-adaptive world-action modeling with
  observation-guided context routing.
\newblock \emph{arXiv preprint arXiv:2606.09811}, 2026.
\newblock URL \url{https://arxiv.org/abs/2606.09811}.

\bibitem[Jitkrittum et~al.(2023)Jitkrittum, Gupta, Menon, Narasimhan, Rawat,
  and Kumar]{jitkrittum2023confidence}
Wittawat Jitkrittum, Neha Gupta, Aditya~K. Menon, Harikrishna Narasimhan, Ankit
  Rawat, and Sanjiv Kumar.
\newblock When does confidence-based cascade deferral suffice?
\newblock In \emph{Advances in Neural Information Processing Systems},
  volume~36, 2023.

\bibitem[Le~Cam and Yang(2000)]{lecam2000asymptotics}
Lucien Le~Cam and Grace~Lo Yang.
\newblock \emph{Asymptotics in Statistics: Some Basic Concepts}.
\newblock Springer, 2 edition, 2000.

\bibitem[Mozannar and Sontag(2020)]{mozannar2020consistent}
Hussein Mozannar and David Sontag.
\newblock Consistent estimators for learning to defer to an expert.
\newblock In \emph{Proceedings of the 37th International Conference on Machine
  Learning}, 2020.

\bibitem[Narasimhan et~al.(2022)]{narasimhan2022posthoc}
Harikrishna Narasimhan et~al.
\newblock Post-hoc estimators for learning to defer to an expert.
\newblock In \emph{Advances in Neural Information Processing Systems}, 2022.

\bibitem[Oquab et~al.(2024)Oquab, Darcet, Moutakanni, Vo, Szafraniec, Khalidov,
  Fernandez, Haziza, Massa, El-Nouby, et~al.]{ocquab2023dinov2}
Maxime Oquab, Timoth{\'e}e Darcet, Th{\'e}o Moutakanni, Huy Vo, Marc
  Szafraniec, Vasil Khalidov, Pierre Fernandez, Daniel Haziza, Francisco Massa,
  Alaa El-Nouby, et~al.
\newblock {DINOv2}: Learning robust visual features without supervision.
\newblock \emph{Transactions on Machine Learning Research}, 2024.

\bibitem[Regol et~al.(2025)Regol, Cotnareanu, Glavas, and
  Coates]{regol2025acquisition}
Florence Regol, Joseph Cotnareanu, Theodore Glavas, and Mark Coates.
\newblock Is the acquisition worth the cost? surrogate losses for consistent
  two-stage classifiers.
\newblock In \emph{Advances in Neural Information Processing Systems},
  volume~38, 2025.

\bibitem[Russell and Wefald(1991)]{russell1991principles}
Stuart Russell and Eric Wefald.
\newblock \emph{Do the Right Thing: Studies in Limited Rationality}.
\newblock MIT Press, 1991.

\bibitem[Seo et~al.(2026)Seo, Kim, and Kwak]{seo2026acid}
Gawon Seo, Dongwon Kim, and Suha Kwak.
\newblock {ACID}: Action consistency via inverse dynamics for planning with
  world models.
\newblock \emph{arXiv preprint arXiv:2607.02403}, 2026.
\newblock URL \url{https://arxiv.org/abs/2607.02403}.

\bibitem[Sivasankar(2026)]{sivasankar2026adaptive}
Achyuthan Sivasankar.
\newblock Adaptive compute in latent world models: When depth helps, hurts, or
  doesn't matter.
\newblock \emph{arXiv preprint arXiv:2607.10203}, 2026.
\newblock URL \url{https://arxiv.org/abs/2607.10203}.

\bibitem[Sun et~al.(2026)Sun, Zhuge, Liu, Gu, Bing, Gan, and
  Tian]{sun2026sants}
Yirui Sun, Guangyu Zhuge, Keliang Liu, Jie Gu, Xinyu Bing, Zhongxue Gan, and
  Chunxu Tian.
\newblock {SANTS}: A state-adaptive scheduler for world action models.
\newblock \emph{arXiv preprint arXiv:2605.27947}, 2026.
\newblock URL \url{https://arxiv.org/abs/2605.27947}.

\bibitem[van Rooyen and Williamson(2014)]{vanrooyen2014lecam}
Brendan van Rooyen and Robert~C. Williamson.
\newblock Le cam meets lecun: Deficiency and generic feature learning.
\newblock \emph{arXiv preprint arXiv:1402.4884}, 2014.
\newblock URL \url{https://arxiv.org/abs/1402.4884}.

\bibitem[Vasilyev et~al.(2026)Vasilyev, Wang, Li, and
  Chen]{vasilyev2026conditional}
Andrey Vasilyev, Yikai Wang, Xiaocheng Li, and Guanting Chen.
\newblock Calibrating conditional risk.
\newblock \emph{arXiv preprint arXiv:2604.20409}, 2026.
\newblock URL \url{https://arxiv.org/abs/2604.20409}.

\bibitem[Wang et~al.(2026)Wang, Zhang, Lin, Luo, Wang, Wang, and
  Qi]{wang2026imagination}
Rui Wang, Yue Zhang, Jiehong Lin, Kuncheng Luo, Jianan Wang, Zhongrui Wang, and
  Xiaojuan Qi.
\newblock When to trust imagination: Adaptive action execution for world action
  models.
\newblock \emph{arXiv preprint arXiv:2605.06222}, 2026.
\newblock URL \url{https://arxiv.org/abs/2605.06222}.

\bibitem[Wilder et~al.(2019)Wilder, Dilkina, and Tambe]{wilder2019melding}
Bryan Wilder, Bistra Dilkina, and Milind Tambe.
\newblock Melding the data-decisions pipeline: Decision-focused learning for
  combinatorial optimization.
\newblock In \emph{Proceedings of the AAAI Conference on Artificial
  Intelligence}, 2019.

\bibitem[Yuan et~al.(2026)Yuan, Dong, Liu, and Zhao]{yuan2026fastwam}
Tianyuan Yuan, Zibin Dong, Yicheng Liu, and Hang Zhao.
\newblock {Fast-WAM}: Do world action models need test-time future imagination?
\newblock \emph{arXiv preprint arXiv:2603.16666}, 2026.
\newblock URL \url{https://arxiv.org/abs/2603.16666}.

\bibitem[Zhao and Ermon(2021)]{zhao2021right}
Shengjia Zhao and Stefano Ermon.
\newblock Right decisions from wrong predictions: A mechanism design
  alternative to individual calibration.
\newblock In \emph{Proceedings of the 24th International Conference on
  Artificial Intelligence and Statistics}, volume 130, pages 2683--2691, 2021.

\bibitem[Zhao et~al.(2021)Zhao, Kim, Sahoo, Ma, and Ermon]{zhao2021calibrating}
Shengjia Zhao, Michael Kim, Roshni Sahoo, Tengyu Ma, and Stefano Ermon.
\newblock Calibrating predictions to decisions: A novel approach to multi-class
  calibration.
\newblock In \emph{Advances in Neural Information Processing Systems}, 2021.

\bibitem[Zhao et~al.(2026)Zhao, Cho, Shen, Zheng, Gao, Cao, and
  Mao]{zhao2026geometric}
Zesen Zhao, Minkyoung Cho, Hui Shen, Boyuan Zheng, Kunxiao Gao, Yulong Cao, and
  Z.~Morley Mao.
\newblock Test-time scaling for world action models via zero-shot geometric
  verification.
\newblock \emph{arXiv preprint arXiv:2607.17454}, 2026.
\newblock URL \url{https://arxiv.org/abs/2607.17454}.

\bibitem[Zhou et~al.(2025)Zhou, Pan, LeCun, and Pinto]{zhou2024dinowm}
Gaoyue Zhou, Hengkai Pan, Yann LeCun, and Lerrel Pinto.
\newblock {DINO-WM}: World models on pre-trained visual features enable
  zero-shot planning.
\newblock In \emph{Proceedings of the 42nd International Conference on Machine
  Learning}, volume 267 of \emph{Proceedings of Machine Learning Research},
  pages 79115--79135, 2025.
\newblock URL \url{https://proceedings.mlr.press/v267/zhou25t.html}.

\end{thebibliography}
\endgroup

\end{document}